# Optimal Excitation Trajectories for System Identification of Underwater Vehicles

Fotis Panetsos and Kostas J. Kyriakopoulos

***Abstract*— In this work, we propose a structured methodology for the system identification of underwater vehicles through the design of optimal excitation trajectories. To this end, the trajectories are parameterized using Bézier curves, which ensure smooth and differentiable motion profiles while facilitating the enforcement of constraints through appropriate manipulation of the control points. An optimization problem is formulated to determine a dynamically feasible excitation trajectory that respects safety limits and maximizes the quality of the collected data, thereby enabling reliable estimation of the vehicle's dynamic parameters using least squares. The proposed methodology is experimentally validated in a laboratory water tank, where the dynamic parameters, identified from the optimized trajectory, are evaluated by predicting the vehicle's velocity through forward simulation on previously unseen trajectories.**

## I. INTRODUCTION

The field of underwater vehicles (UVs) has seen significant development in recent years due to their versatility and applicability to a wide range of tasks, including aquaculture, oceanographic research, environmental monitoring, archaeological surveying, and offshore infrastructure inspection [1], [2], [3], [4]. However, the complexity and challenges associated with such missions highlight the need to enhance the autonomy and reliability of UVs. In this context, having an accurate dynamic model can play a key role. Such models are essential for designing advanced model-based control schemes that outperform traditional ones. Furthermore, the ability to predict the vehicle's velocity using the identified dynamics enables the implementation of model-based state estimation techniques [5], online fault diagnosis and fault-tolerant control [6], as well as disturbance estimation methods [7].

A common approach to obtaining accurate dynamic models of UVs involves conducting captive model tests in towing tanks using planar motion mechanisms [8]. However, these procedures are time-consuming and require access to expensive and specialized facilities. Computational fluid dynamics (CFD) simulations offer an alternative; nevertheless, their accuracy relies not only on detailed computer-aided design (CAD) models of the UV, but also on precise knowledge of inertial properties, i.e., mass distribution and moments of inertia, which are often difficult to obtain [9]. Analytical and semi-empirical estimation methods may also be employed, particularly for UVs with simple geometries such as torpedo-shaped designs. However, even in these cases, the estimated parameters should be adjusted based on experimental data [10]. Consequently, system identification (SI) techniques have garnered increasing attention due to their practicability and time efficiency as they rely exclusively on data acquired from onboard sensors during field experiments and can thus be readily repeated. This is particularly advantageous for modular UVs, which accommodate varying task-specific payloads and sensors that alter the vehicle's dynamics.

The authors are with the Center for AI & Robotics (CAIR) and the Electrical Eng. Program, Engineering Division, New York University Abu Dhabi, Abu Dhabi, United Arab Emirates. E-mails: {f.panetsos, kkyria}@nyu.edu

### A. Related Works

SI techniques for UVs can be broadly categorized into adaptive identification algorithms, Kalman filter-based methods, and Least Squares (LS) approaches. The aforementioned techniques mainly rely on the well-established equations of motion for UVs [11], exploiting their linearity w.r.t. the dynamic parameters. In addition to these gray-box SI methods, black-box machine learning (ML) approaches have also been explored [12], [13]. Although data-driven methods offer a promising alternative, inherent challenges, such as hyperparameter tuning, overfitting, and lack of interpretability, hinder their deployment in safety-critical underwater applications.

In the context of adaptive identification algorithms, [14] addresses the online estimation of dynamic model parameters for fully coupled 6-DoF UVs. This adaptive approach is supported by a theoretical stability analysis and does not rely on acceleration measurements or a simultaneous model-based trajectory tracking controller. In [15], the authors investigate the simultaneous estimation of plant-model and actuator parameters for underactuated UVs, such as torpedo-shaped vehicles. The proposed adaptive scheme is theoretically proven to converge to the true set of parameters under arbitrary control laws, provided that a persistence of excitation condition is satisfied.

Kalman filter-based techniques have also been exploited for joint state and parameter estimation. In [16], an asynchronous Modified Dual Unscented Kalman Filter is used to estimate, in parallel and online, both the state and the dynamic parameters of a 3-DoF UV. Moreover, [17] evaluates the performance of the Extended Kalman Filter (EKF), Cubature Kalman Filter (CKF), and Transformed Unscented Kalman Filter (TUKF) for identifying the hydrodynamic coefficients of a 6-DoF UV in simulation. Nevertheless, despite the demonstrated capabilities of both adaptive identification algorithms and Kalman filter-based methods, the selection of gains and parameter initialization remain critical open issues.

As a result, the majority of the works in the literature adopt LS techniques to estimate the dynamic parameters of UVs, due to their simplicity. In [18], a two-step identification procedure is presented to estimate a lumped parameter model under the assumption of decoupled motion, while accounting for propeller–hull and propeller–propeller interactions. Constant and sinusoidal inputs are applied to estimate drag and inertial parameters, respectively. In [19], a comparative study is conducted between two LS methods for the identification of the parameters governing the surge, yaw, and pitch motions. Each of these DoFs is independently excited using step and PRBS input signals. A hybrid estimation strategy is introduced in [9] to identify the hydrodynamic parameters affecting horizontal motion, where an initial estimate is obtained via the iteratively recursive LS method and subsequently refined using the improved particle swarm optimization algorithm. In [20], an unsupervised and online SI framework is proposed for a 4-DoF UV based on recursive LS. The UV is excited through open-loop periodic and step inputs, while a collision avoidance strategy is incorporated to ensure safe operation within a constrained tank. Finally, [21] evaluates the accuracy of various 6-DoF coupled dynamic models. Both ordinary and total LS methods are explored to identify the unknown dynamic parameters, with the translational DoFs excited by tracking sinusoidal position and velocity reference trajectories under closed-loop control, while the rotational DoFs are excited using open-loop sinusoidal inputs.

Despite the extensive literature on the SI of UVs, a critical but often neglected point of discussion concerns the excitation of the system. Most existing works either depend on large datasets, collected through extended field trials, or employ basic inputs, such as step and sinusoidal signals with heuristically selected parameters, typically applied independently to each DoF. Furthermore, the aforementioned approaches typically rely on simplifying assumptions, such as known center of mass (CoM) or fully decoupled motion. As a result, the validity of the identified models is often limited to basic maneuvers, thereby failing to capture the agile and coupled dynamic behavior of modern UVs. The generation of excitation trajectories has been extensively explored in the literature for other robotic platforms, particularly humanoid robots [22] and manipulators [23], [24], [25], [26]. These methods typically entail optimizing the parameters of excitation trajectories, commonly represented using Fourier series or B-splines, by minimizing the condition number of the regressor matrix. The resulting trajectories are then employed in conjunction with LS techniques to identify the dynamic parameters. However, such systematic approaches have not been adopted in the context of UVs.

### B. Contributions

In this work, we introduce a systematic framework for the identification of the dynamic parameters governing the coupled 4-DoF motion of a UV, specifically the surge, sway, heave, and yaw dynamics. The proposed methodology lies in designing optimal excitation trajectories using Bézier curves, which can represent a wide variety of smooth motion profiles while also inherently supporting constraint enforcement through appropriate handling of the control points. To this end, a feasible optimization problem is formulated to find the set of control points that minimizes the condition number of the regressor matrix, thereby ensuring a well-conditioned LS solution for parameter identification, while simultaneously satisfying safety and operational constraints. In contrast to prior approaches that employ large datasets, basic input signals, or simplifying assumptions such as decoupled motion, our framework provides a structured approach for generating excitation trajectories tailored to UVs, enabling reliable identification of their dynamic parameters.

The efficacy of the proposed methodology is experimentally verified in a laboratory water tank using the VideoRay Defender vehicle. Initially, the UV tracks the optimized trajectory and its dynamic parameters are identified through LS. Subsequently, the accuracy of the identified parameters is evaluated in cross-validation, where the vehicle's velocity is predicted through forward simulation of the dynamic model and compared against the measured data.

## II. Equations of Motion

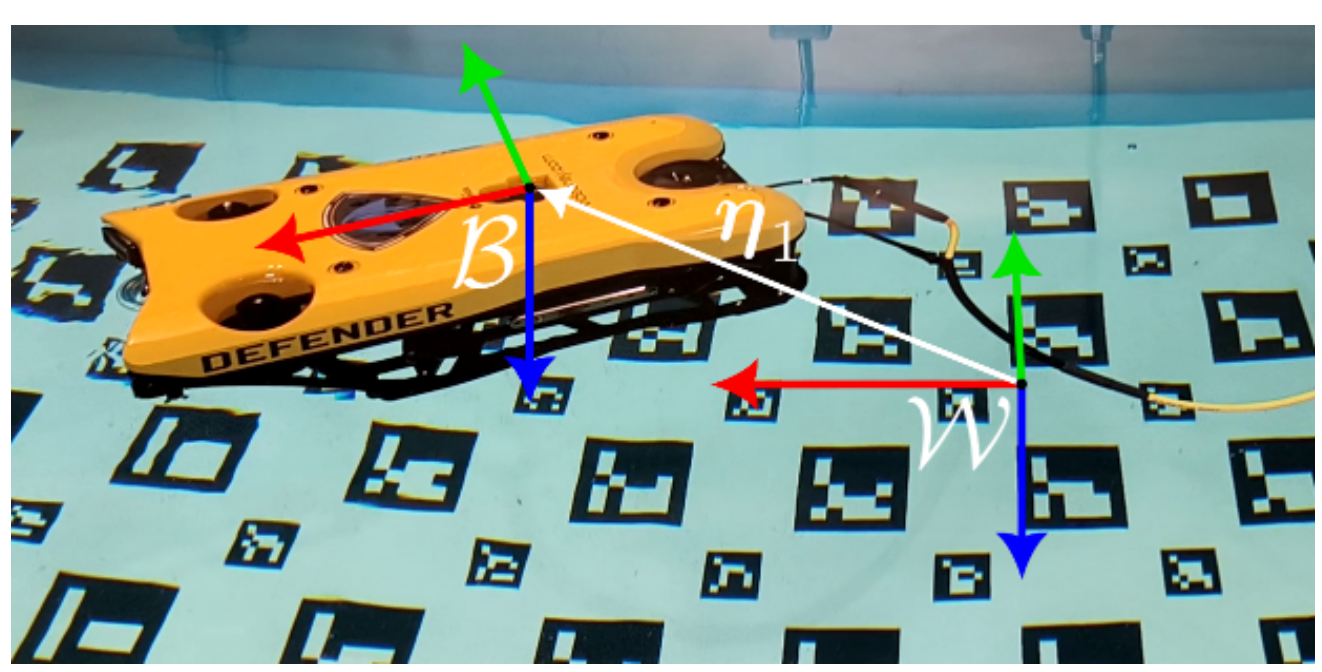

Fig. 1. The VideoRay Defender vehicle in the laboratory water tank. The world frame $\mathcal{W}$, the body-fixed frame $\mathcal{B}$, and the vehicle's position $\boldsymbol{\eta}_1$ are depicted. The $x$, $y$, and $z$ axes of each frame are represented by red, green, and blue arrows respectively.

In this section, we present the dynamic model that governs the 4-DoF motion of a UV according to [11].

Consider the UV, specifically the VideoRay Defender used in this study, operating within a water tank, and let $\mathcal{W}$ and $\mathcal{B}$ denote the world and body-fixed frames, respectively, as illustrated in Fig. 1. The origin of $\mathcal{W}$ is located at the water surface of the tank. Since the exact location of the vehicle's CoM is unknown, the origin of $\mathcal{B}$ is chosen to coincide with the optical center of the onboard downward-looking camera, which is used to estimate the vehicle's position, as detailed in Section IV-A. The position of the UV is denoted by $\boldsymbol{\eta}_1 = \begin{bmatrix} x & y & z \end{bmatrix}^T$, while its orientation is parameterized by the Euler roll–pitch–yaw ($\phi$–$\theta$–$\psi$) angles. In this work, our objective is to identify the dynamics associated with four DoFs of the vehicle, namely the surge, sway, heave, and yaw motions. To this end, by neglecting the roll and pitch dynamics and assuming that the respective angles and angular velocities are close to zero, the pose of the UV is expressed as $\boldsymbol{\eta} = \begin{bmatrix} \boldsymbol{\eta}_1^T & \psi \end{bmatrix}^T \in \mathbb{R}^4$. The time derivative of the pose is related to the vehicle's body-fixed velocity by:

$$\dot{\boldsymbol{\eta}} = \mathbf{J}(\boldsymbol{\eta})\mathbf{v}, \tag{1}$$

where $\mathbf{J}(\boldsymbol{\eta})$ is the Jacobian matrix transforming velocities from $\mathcal{B}$ to $\mathcal{W}$, and $\mathbf{v} = \begin{bmatrix} u & v & w & r \end{bmatrix}^T$ is the vehicle's body-fixed velocity, with $u$, $v$, and $w$ denoting the linear velocities in surge, sway, and heave, respectively, and $r$ representing the yaw rate.

The dynamics of the UV can be described as follows:

$$\mathbf{M}\dot{\mathbf{v}} + \mathbf{C}(\mathbf{v})\mathbf{v} + \mathbf{D}\mathbf{v} + \mathbf{g} = \boldsymbol{\tau}, \tag{2}$$

where $\mathbf{M} \in \mathbb{R}^{4\times4}$ is the inertia matrix, incorporating both the rigid-body and the added mass terms; $\mathbf{C}(\mathbf{v}) \in \mathbb{R}^{4\times4}$ is the Coriolis-centripetal matrix, including added mass effects; $\mathbf{D} \in \mathbb{R}^{4\times4}$ is the hydrodynamic damping matrix; $\mathbf{g} \in \mathbb{R}^4$ represents the restoring forces due to gravity and buoyancy; and $\boldsymbol{\tau} \in \mathbb{R}^4$ is the control input vector, comprising forces and yaw torque applied to the UV w.r.t. $\mathcal{B}$.

Assuming that the vehicle operates in an ideal fluid, the inertia matrix $\mathbf{M}$ is symmetric and positive definite, i.e., $\mathbf{M} = \mathbf{M}^T > \mathbf{0}$. Furthermore, due to the port/starboard symmetry of the UV (symmetry w.r.t. the $xz$-plane), the inertia matrix can be expressed as:

$$\mathbf{M} = \begin{bmatrix} m_{11} & 0 & m_{13} & 0 \\ 0 & m_{22} & 0 & m_{26} \\ m_{13} & 0 & m_{33} & 0 \\ 0 & m_{26} & 0 & m_{66} \end{bmatrix}. \tag{3}$$

It should be noted that the origin of $\mathcal{B}$ lies on the $xz$-plane of symmetry. The Coriolis-centripetal matrix $\mathbf{C}(\mathbf{v})$ is derived directly from the inertia matrix $\mathbf{M}$, as described in [11].

Furthermore, under the assumption that the vehicle moves at low speed, the linear damping terms are dominant relative to the nonlinear quadratic contributions. Consequently, the damping matrix $\mathbf{D}$ can be represented as a constant $4 \times 4$ matrix with 16 fixed elements. Finally, considering negligible roll and pitch angles, the restoring forces are given by $\mathbf{g} = \begin{bmatrix} 0 & 0 & -(W-B) & 0 \end{bmatrix}^T$, where $W$ and $B$ denote the weight and buoyancy of the UV, correspondingly.

### A. Identification Model

By leveraging the linearity of the dynamic model of the UV w.r.t. its parameters, Eq. 2 can be rewritten as follows:

$$\boldsymbol{\tau} = \mathbf{Y}\left(\dot{\mathbf{v}}, \mathbf{v}\right)\boldsymbol{\pi}, \tag{4}$$

where $\mathbf{Y}\left(\dot{\mathbf{v}}, \mathbf{v}\right) \in \mathbb{R}^{4\times n_p}$ is the regressor matrix and $\boldsymbol{\pi} \in \mathbb{R}^{n_p}$ denotes the base parameters, i.e., the minimum set of parameters required to fully describe the dynamic behavior of the UV. In this case, the base parameters $\boldsymbol{\pi}$ consist of: (i) the unique nonzero entries of the symmetric inertia matrix $\mathbf{M}$ (Eq. 3), namely $\{m_{11},\ m_{22},\ m_{33},\ m_{66},\ m_{13},\ m_{26}\}$; (ii) the 16 entries of the damping matrix $\mathbf{D}$; and (iii) the difference between the vehicle's weight and buoyancy $(W - B)$.

### B. Actuation Model

The VideoRay Defender is equipped with seven identical thrusters, specifically four horizontal thrusters responsible for surge, sway, and yaw maneuvers, and three vertical thrusters that control heave, roll, and pitch motions. The resulting forces and moments $\boldsymbol{\tau}_\upsilon \in \mathbb{R}^6$ (including roll and pitch moments) applied to the vehicle are calculated as:

$$\boldsymbol{\tau}_\upsilon = \mathbf{B}_\upsilon \mathbf{f}, \tag{5}$$

where $\mathbf{f} = \begin{bmatrix} f_1, \cdots, f_7 \end{bmatrix}^T \in \mathbb{R}^7$ is the vector of thruster forces, and $\mathbf{B}_\upsilon \in \mathbb{R}^{6\times7}$ denotes the thruster allocation matrix, given by:

$$\mathbf{B}_\upsilon = \begin{bmatrix} \mathbf{e}_1 & \cdots & \mathbf{e}_7 \\ \mathbf{r}_1 \times \mathbf{e}_1 & \cdots & \mathbf{r}_7 \times \mathbf{e}_7 \end{bmatrix}, \tag{6}$$

where $\mathbf{e}_i \in \mathbb{R}^3$ is the unit vector that defines the direction of the $i$-th thruster w.r.t. $\mathcal{B}$, and $\mathbf{r}_i \in \mathbb{R}^3$ denotes its position vector relative to $\mathcal{B}$.

In practice, each thruster of the UV is actuated via a normalized control input $\upsilon_i \in [-1, 1]$, corresponding to the duty cycle of a Pulse Width Modulation (PWM) signal. Consequently, a mapping between the control input $\upsilon_i$ and the generated thruster force $f_i$ is required to identify the vehicle's dynamics. To this end, two distinct fifth-order polynomial models - for forward and reverse rotation - are derived from experimental data collected through static tests on a single thruster, while also incorporating a narrow dead-band zone around zero input. It should be highlighted that the thruster's transient response is assumed to be negligible compared to the vehicle's dynamics. Although this assumption may not hold under high-frequency or abrupt signals, it remains reasonable in the context of this work, where the UV is commanded to follow smooth trajectories.

## III. Excitation Trajectory

To identify the dynamic parameters $\boldsymbol{\pi}$ of the UV, the regressor matrices and the control inputs of Eq. 4 should be collected at multiple time instants $t_k,\ k = 1, \cdots, N$:

$$\bar{\boldsymbol{\tau}} = \begin{bmatrix} \boldsymbol{\tau}(t_1) \\ \vdots \\ \boldsymbol{\tau}(t_N) \end{bmatrix}, \bar{\mathbf{Y}} = \begin{bmatrix} \mathbf{Y}\left(\dot{\mathbf{v}}(t_1), \mathbf{v}(t_1)\right) \\ \vdots \\ \mathbf{Y}\left(\dot{\mathbf{v}}(t_N), \mathbf{v}(t_N)\right) \end{bmatrix}. \tag{7}$$

Given a sufficiently rich dataset, the parameters $\boldsymbol{\pi}$ can be estimated using an LS method. However, the accuracy of the LS solution is highly sensitive to the numerical properties of the regressor matrix $\bar{\mathbf{Y}}$. Among various criteria used to evaluate the quality of $\bar{\mathbf{Y}}$, the 2-norm condition number is one of the most commonly employed. Since $\bar{\mathbf{Y}}$ is determined by the trajectory of the UV, the objective lies in designing an excitation trajectory, i.e., $\boldsymbol{\eta}_r, \mathbf{v}_r, \dot{\mathbf{v}}_r$, that minimizes the condition number of $\bar{\mathbf{Y}}$. A direct solution to this optimization problem results in a series of discrete trajectory points, which should then be interpolated to obtain a continuous and feasible trajectory. To address this limitation and reduce computational complexity, parameterized trajectories are typically employed, and the problem is reformulated as an optimization over a set of trajectory parameters.

### A. Trajectory Parameterization

In this work, the excitation trajectory is parameterized using Bézier curves, which can model smooth, continuous, and differentiable trajectories. In contrast to Fourier series - commonly employed in the literature but restricted to periodic profiles - Bézier curves are capable of representing a wider range of trajectories, thereby offering greater flexibility. Furthermore, due to their inherent properties, as discussed in Section III-B, trajectory constraints can be enforced directly by manipulating the control points. This eliminates the need to impose constraints at each trajectory sample, as required when using Fourier series, thereby significantly reducing the computational cost of the optimization problem.

More specifically, we consider a multi-segment excitation trajectory $\boldsymbol{\eta}_r$ for the vehicle's pose, defined as:

$$\boldsymbol{\eta}_r(t) = \begin{cases} \boldsymbol{\eta}_r^1(t), & t \in [T_0, T_1], \\ \vdots \\ \boldsymbol{\eta}_r^m(t), & t \in [T_{m-1}, T_m], \end{cases} \tag{8}$$

where $m$ denotes the number of segments, $\boldsymbol{\eta}_r^j(t) \in \mathbb{R}^4$ is the $j$-th segment of the trajectory with $j = 1, \cdots, m$, and $[T_{j-1}, T_j]$ is the time interval of each segment. It should be noted that a multi-segment trajectory is adopted to allow the enforcement of distinct constraints on individual segments.

Each segment $\boldsymbol{\eta}_r^j$ of the trajectory is represented by a Bézier curve of order $n$ according to the following equation:

$$\boldsymbol{\eta}_r^j(t) = \sum_{i=0}^{n} b_{i,n}(\tau)\mathbf{c}_i^j, \tag{9}$$

where $\mathbf{c}_i^j = \begin{bmatrix} c_{x,i}^j & c_{y,i}^j & c_{z,i}^j & c_{\psi,i}^j \end{bmatrix}^T \in \mathbb{R}^4$, are the $n+1$ control points associated with the $j$-th segment, $\tau \in [0, 1]$ is a normalized time parameter, and $b_{i,n}(\tau)$ denotes the $i$-th Bernstein basis polynomial of degree $n$. The parameter $\tau$ is linearly mapped to the actual time interval of the segment $[T_{j-1}, T_j]$. Consequently, each segment is fully characterized by the set of control points $\mathbf{c}^j = \{\mathbf{c}_0^j, \cdots, \mathbf{c}_n^j\} \in \mathbb{R}^{4(n+1)}$, while the complete multi-segment excitation trajectory is described by the concatenated set $\mathbf{c} = \{\mathbf{c}^1, \cdots, \mathbf{c}^m\}$, comprising a total of $4m(n+1)$ parameters.

By successively differentiating the Bézier curves, the velocity $\dot{\boldsymbol{\eta}}_r$ and acceleration $\ddot{\boldsymbol{\eta}}_r$ of the vehicle, expressed in $\mathcal{W}$, can be computed. The corresponding body-fixed velocity $\mathbf{v}_r$ is then given by:

$$\mathbf{v}_r = \mathbf{J}^{-1}(\boldsymbol{\eta}_r)\,\dot{\boldsymbol{\eta}}_r, \tag{10}$$

while the body-fixed acceleration $\dot{\mathbf{v}}_r$ is subsequently obtained by differentiating Eq. 10.

### B. Constraints

To generate the aforementioned multi-segment trajectory, it is essential to incorporate constraints into the optimization problem to ensure both dynamical feasibility and operational safety. To this end, the following constraints are imposed:

*1) Pose Constraints:* By leveraging the convex hull property of Bézier curves, which states that the curve lies entirely within the convex hull formed by its control points, the following constraints are formulated for the $j$-th segment:

$$\boldsymbol{\eta}_{min}^j \le \mathbf{c}_i^j \le \boldsymbol{\eta}_{max}^j, \tag{11}$$

where $\boldsymbol{\eta}_{min}^j, \boldsymbol{\eta}_{max}^j \in \mathbb{R}^4$ are the pose limits. For instance, these constraints include bounds on the vehicle's position to ensure that it remains within the boundaries of the tank.

*2) Velocity and Acceleration Constraints:* To avoid dynamically infeasible maneuvers, constraints are applied to the vehicle's velocity and acceleration. For this purpose, we exploit another fundamental property of Bézier curves, according to which the first derivative of a Bézier curve of order $n$, defined by control points $\mathbf{c}_i^j$, is itself a Bézier curve of order $n-1$ with control points given by $n(\mathbf{c}_i^j - \mathbf{c}_{i-1}^j)$. Thus, by again utilizing the convex hull property, the following constraints are considered:

$$\begin{aligned} n\left(\mathbf{c}_i^j - \mathbf{c}_{i-1}^j\right)/\left(T_j - T_{j-1}\right) &\in [\dot{\boldsymbol{\eta}}_{min}^j, \dot{\boldsymbol{\eta}}_{max}^j], \\ \frac{n(n-1)\left(\mathbf{c}_i^j - 2\mathbf{c}_{i-1}^j + \mathbf{c}_{i-2}^j\right)}{\left(T_j - T_{j-1}\right)^2} &\in [\ddot{\boldsymbol{\eta}}_{min}^j, \ddot{\boldsymbol{\eta}}_{max}^j], \end{aligned} \tag{12}$$

where $\dot{\boldsymbol{\eta}}_{min}^j$, $\dot{\boldsymbol{\eta}}_{max}^j \in \mathbb{R}^4$ and $\ddot{\boldsymbol{\eta}}_{min}^j$, $\ddot{\boldsymbol{\eta}}_{max}^j \in \mathbb{R}^4$ represent the velocity and acceleration bounds for the $j$-th segment of the trajectory. It is highlighted that these constraints are enforced w.r.t. $\mathcal{W}$, rather than the body-fixed frame $\mathcal{B}$, in order to reduce the computational complexity of the optimization problem. If the velocity and acceleration constraints were expressed in $\mathcal{B}$, the resulting constraints would involve nonlinear terms and would need to be evaluated at each trajectory sample, in contrast to the $4mn$ and $4m(n-1)$ linear constraints of Eq. 12.

*3) $C^2$ Continuity Constraints*: To ensure smooth transitions between consecutive segments, we leverage the endpoint interpolation property of Bézier curves, which means that the curve passes through its first and last control points. Specifically, $C^2$ continuity is enforced through the following equality constraints between the $j$-th and $(j+1)$-th segments:

$$\begin{gathered} \mathbf{c}_n^j = \mathbf{c}_0^{j+1} \\ \frac{\mathbf{c}_n^j - \mathbf{c}_{n-1}^j}{(T_j - T_{j-1})} = \frac{\mathbf{c}_1^{j+1} - \mathbf{c}_0^{j+1}}{(T_{j+1} - T_j)} \\ \frac{\mathbf{c}_n^j - 2\mathbf{c}_{n-1}^j + \mathbf{c}_{n-2}^j}{(T_j - T_{j-1})^2} = \frac{\mathbf{c}_2^{j+1} - 2\mathbf{c}_1^{j+1} + \mathbf{c}_0^{j+1}}{(T_{j+1} - T_j)^2}. \end{gathered} \tag{13}$$

*4) Initial and Terminal Constraints*: Finally, constraints are imposed to ensure that the trajectory starts and ends at rest, i.e., with zero velocity and acceleration at the initial and final time instants, and that the pose begins at the specified initial configuration $\boldsymbol{\eta}_0$ and reaches the desired final one $\boldsymbol{\eta}_f$:

$$\begin{gathered} \mathbf{c}_0^1 = \boldsymbol{\eta}_0, \ \mathbf{c}_n^m = \boldsymbol{\eta}_f \\ \mathbf{c}_1^1 - \mathbf{c}_0^1 = \mathbf{c}_n^m - \mathbf{c}_{n-1}^m = \mathbf{0} \\ \mathbf{c}_2^1 - 2\mathbf{c}_1^1 + \mathbf{c}_0^1 = \mathbf{c}_n^m - 2\mathbf{c}_{n-1}^m + \mathbf{c}_{n-2}^m = \mathbf{0}. \end{gathered} \tag{14}$$

### C. Optimization Problem

The excitation trajectory is ultimately obtained by solving the following nonlinear optimization problem, subject to linear equality and inequality constraints:

$$\begin{aligned} &\min_{\mathbf{c}} \operatorname{cond}\left(\bar{\mathbf{Y}}\left(\dot{\mathbf{v}}_r, \mathbf{v}_r, \boldsymbol{\eta}_r\right)\right) \\ &\text{s.t.} \quad \text{Eq. } 11,\ 12,\ 13,\ 14. \end{aligned} \tag{15}$$

The trajectory spans a time interval of 280s, which is divided into $m = 5$ segments of equal duration, each represented by a Bézier curve of order $n = 20$. In the first four segments, constraints are designed such that a single DoF is predominantly excited in sequence (surge, sway, heave, yaw), while motion in the remaining DoFs is more tightly constrained. For instance, in the first segment, surge is primarily excited by relaxing the translational constraints along the $x$-axis while keeping the yaw angle near zero, so that motion along the $x$-axis of the world frame $\mathcal{W}$ corresponds closely to surge. Simultaneously, constraints associated with the other motions are maintained near zero. This design choice facilitates the identification of the dominant diagonal dynamic parameters for each DoF. In the final segment, all constraints are relaxed to enable simultaneous excitation of all DoFs and the identification of coupled dynamic terms.

To solve the optimization problem, the trajectory is discretized with a fixed time step of 0.2s. Symbolic expressions for both the excitation trajectory $\boldsymbol{\eta}_r$ and the regressor matrix $\bar{\mathbf{Y}}$ are constructed using CasADi [27]. The resulting nonlinear programming problem is solved using the *fmincon* function in MATLAB with the Sequential Quadratic Programming (SQP) algorithm. The average computational time required to obtain a solution is 74.67s on a laptop with the Windows operating system, an Intel i7 CPU, and 16 GB of RAM. However, due to the complexity and non-convex nature of the problem, this approach does not guarantee convergence to the global minimum. As a trade-off between computational cost and solution quality, the optimization is initialized from multiple starting points within the bounds defined by Eq. 11. Eventually, by following this procedure, the regressor matrix of the optimized excitation trajectory exhibits a condition number of 160.10, as shown in Fig. 2.

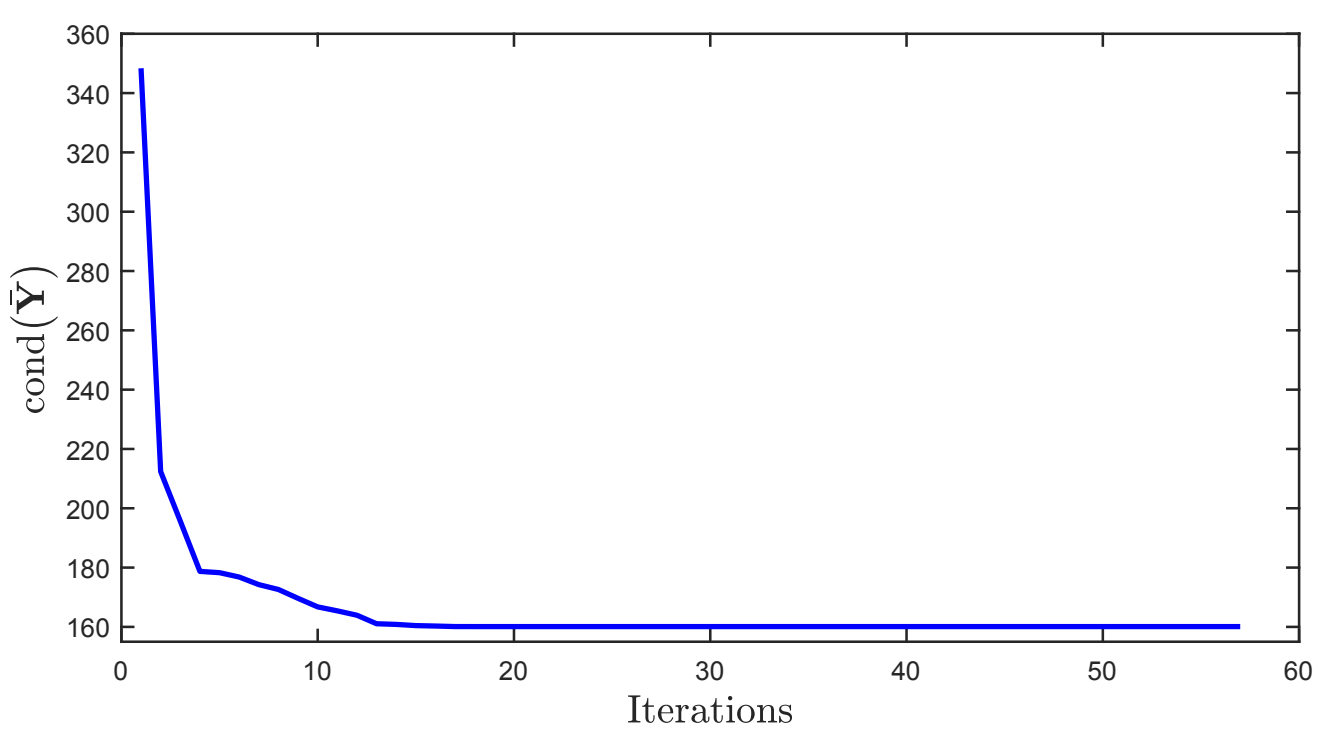

Fig. 2. The evolution of the condition number of the regressor matrix during the optimization process.

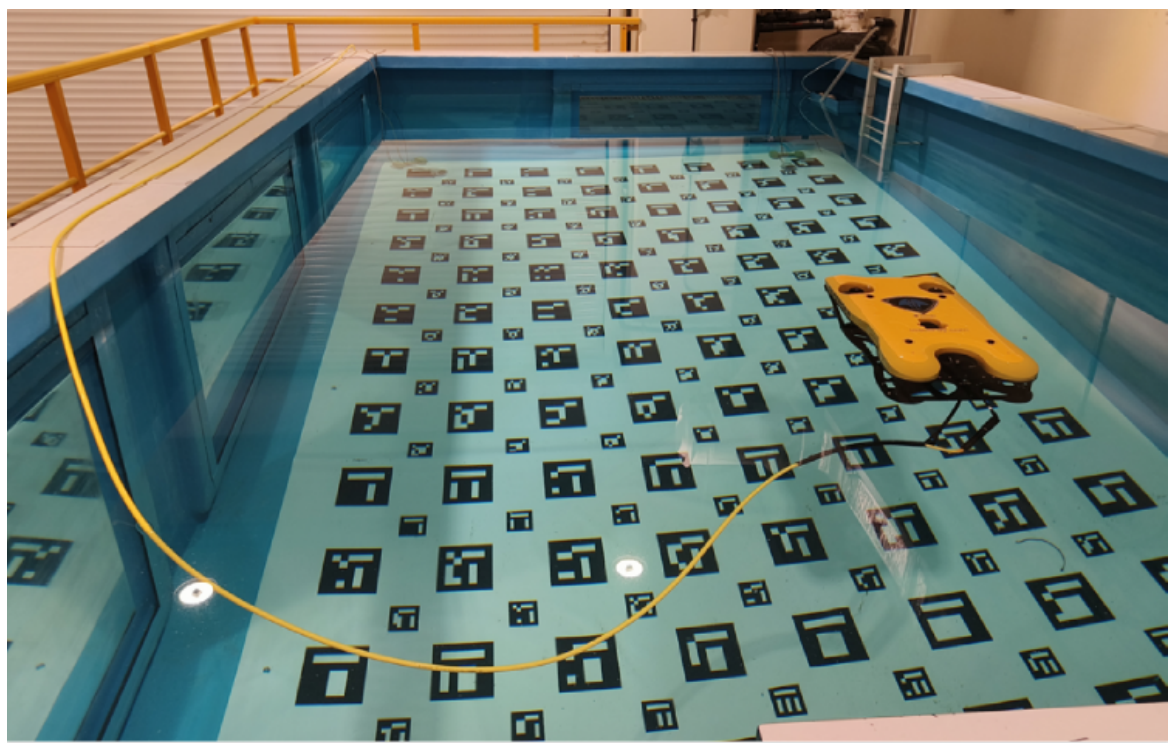

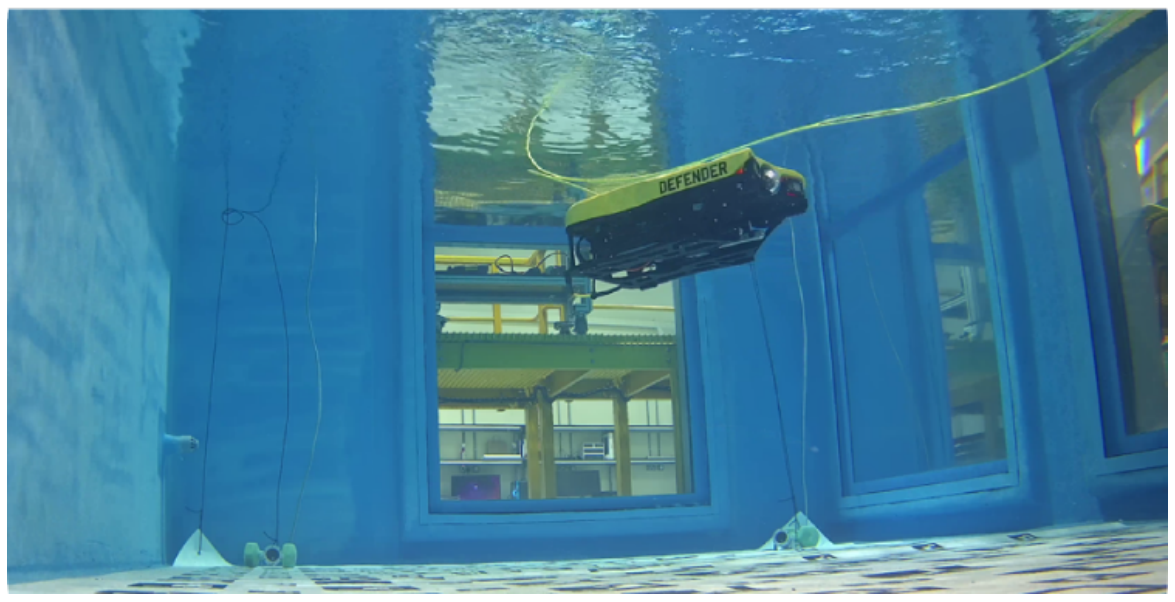

Fig. 3. External and internal views of the laboratory water tank.

## IV. Experimental Results

### A. Experimental Setup

The experiments were conducted in a laboratory water tank with dimensions 6.3m × 3.7m × 1.8m. As depicted in Fig. 3, an ArUco board [28], consisting of markers of two different sizes to facilitate detection at varying depths, was placed at the bottom of the pool. The world frame $\mathcal{W}$ is defined with its origin at the water surface directly above the center of the board, and its $x$-axis aligned along the length of the board.

Regarding the VideoRay Defender used in this study, the vehicle is equipped with a depth sensor, a downward-looking camera employed to detect the ArUco board and estimate the vehicle's position, and an AHRS that provides its orientation w.r.t. the North-East-Down (NED) frame, as well as its angular velocity. It should be mentioned that the orientation measurements from the AHRS, relative to the NED frame, are aligned with the world frame $\mathcal{W}$ through a constant rotation matrix. The measurements obtained by the detection algorithm, the depth sensor, and the AHRS are fused with the aid of a Kalman Filter to estimate the state of the UV. Bidirectional communication with the vehicle's sensors and thrusters is established through the tether. All the software required for communication, state estimation, and control of the vehicle is implemented in C++ and Python, based on the Robot Operating System (ROS) [29].

### B. System Identification

Initially, the dynamic parameters of the vehicle associated with the surge, sway, heave, and yaw motions were identified according to the proposed methodology. More precisely, a PID controller was employed to track the optimized trajectory while simultaneously maintaining the roll and pitch angles, as well as the corresponding angular velocities, close to zero. It should be highlighted that the tracking accuracy of the PID controller directly affects the identification procedure. More specifically, tracking errors must remain sufficiently small so that the resulting condition number of the regressor matrix remains close to the one obtained during the optimization process. To this end, the PID controller is fine-tuned to satisfactorily track the excitation trajectory in the laboratory water tank. The excitation and actual trajectories, along with the pose constraints enforced during the optimization, are illustrated in Fig. 4. As explained in Section III-C, each DoF was excited consecutively over the first four segments, while in the final segment all constraints were relaxed to their maximum allowable values. Based on the data recorded during the aforementioned experiment, the dynamic parameters of the vehicle were estimated using the LS method. To estimate the acceleration, cubic spline interpolation was applied to the velocity measurements, and the acceleration was subsequently obtained by analytically differentiating the fitted splines. This approach resulted in a substantially less noisy estimate compared to direct numerical differentiation of the discrete velocity measurements.

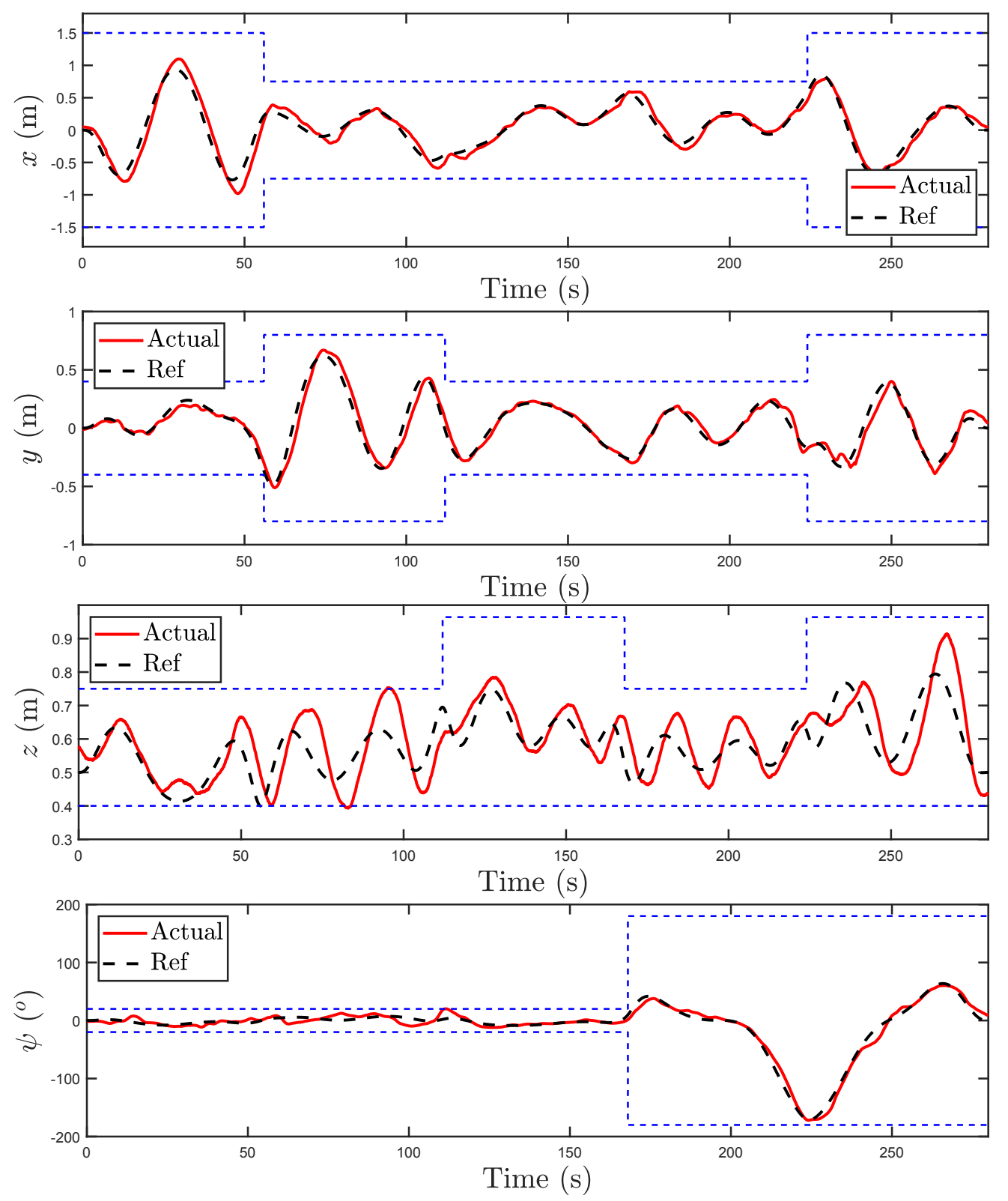

Fig. 4. The excitation trajectory $\boldsymbol{\eta}_r$ and the actual pose of the vehicle $\boldsymbol{\eta}$. The corresponding pose limits for each segment of the excitation trajectory are indicated by dashed blue lines.

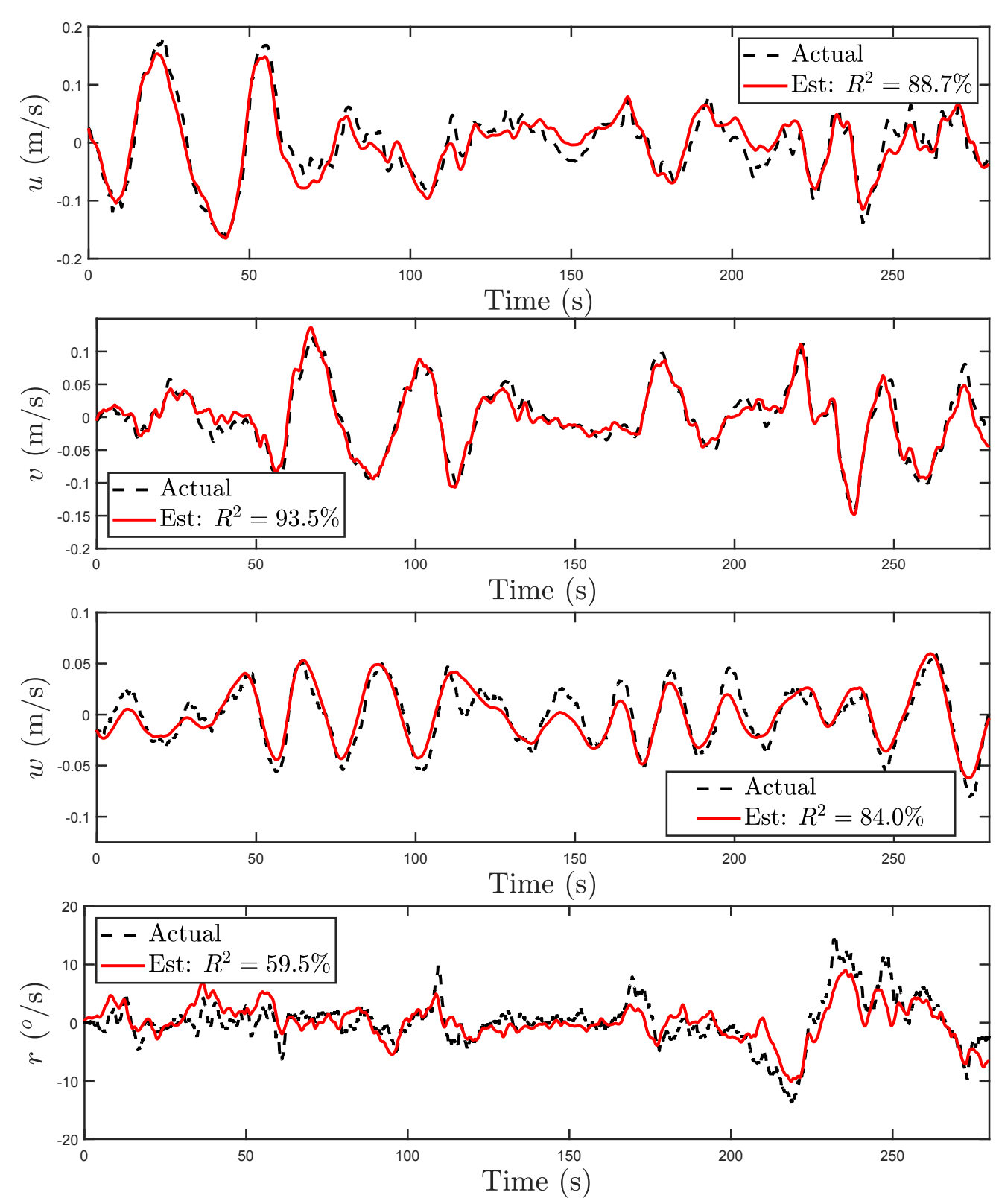

Fig. 5. Actual and simulated body velocities based on the estimated model, with corresponding $R^2$ values for each DoF, during the identification experiment.

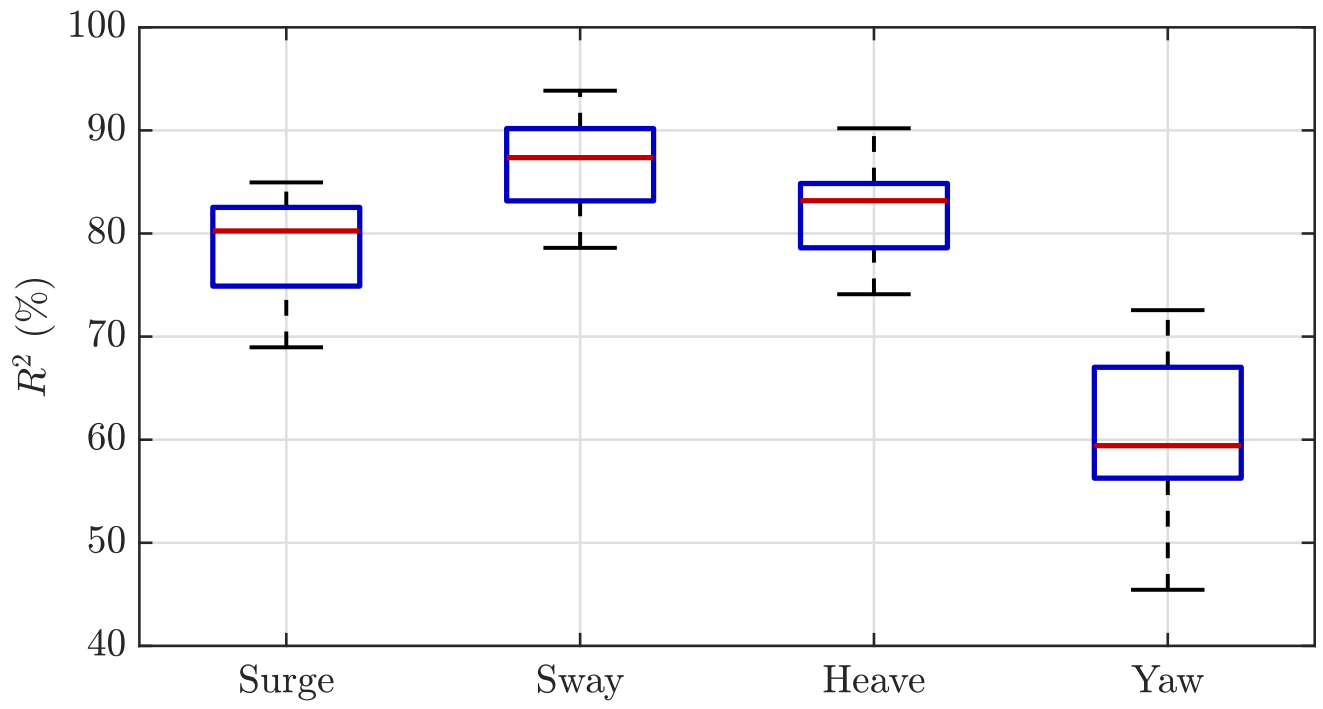

Fig. 6. Box plot illustrating the $R^2$ values between the simulated and actual body velocities for each DoF across 15 experiments. The median, interquartile range, minimum, and maximum values are shown.

To evaluate the accuracy of the identified parameters, the dynamic model was propagated forward in simulation given the control inputs, and the estimated body velocities were compared with the velocity measurements recorded during the experiment via the $R^2$ metric, as depicted in Fig. 5. The discrepancies between the actual and simulated velocities can be attributed to several factors, including interactions among the propellers and between the propellers and the hull, which reduce the efficiency of the thrusters compared to the nominal thrust curve, the presence of the tether, and other unmodeled effects. In particular, the tether significantly affects the yaw dynamics, resulting in a lower $R^2$ value compared to the other DoFs. Nevertheless, the results demonstrate that the identified model can reliably capture the system dynamics and successfully predict the vehicle's body velocities.

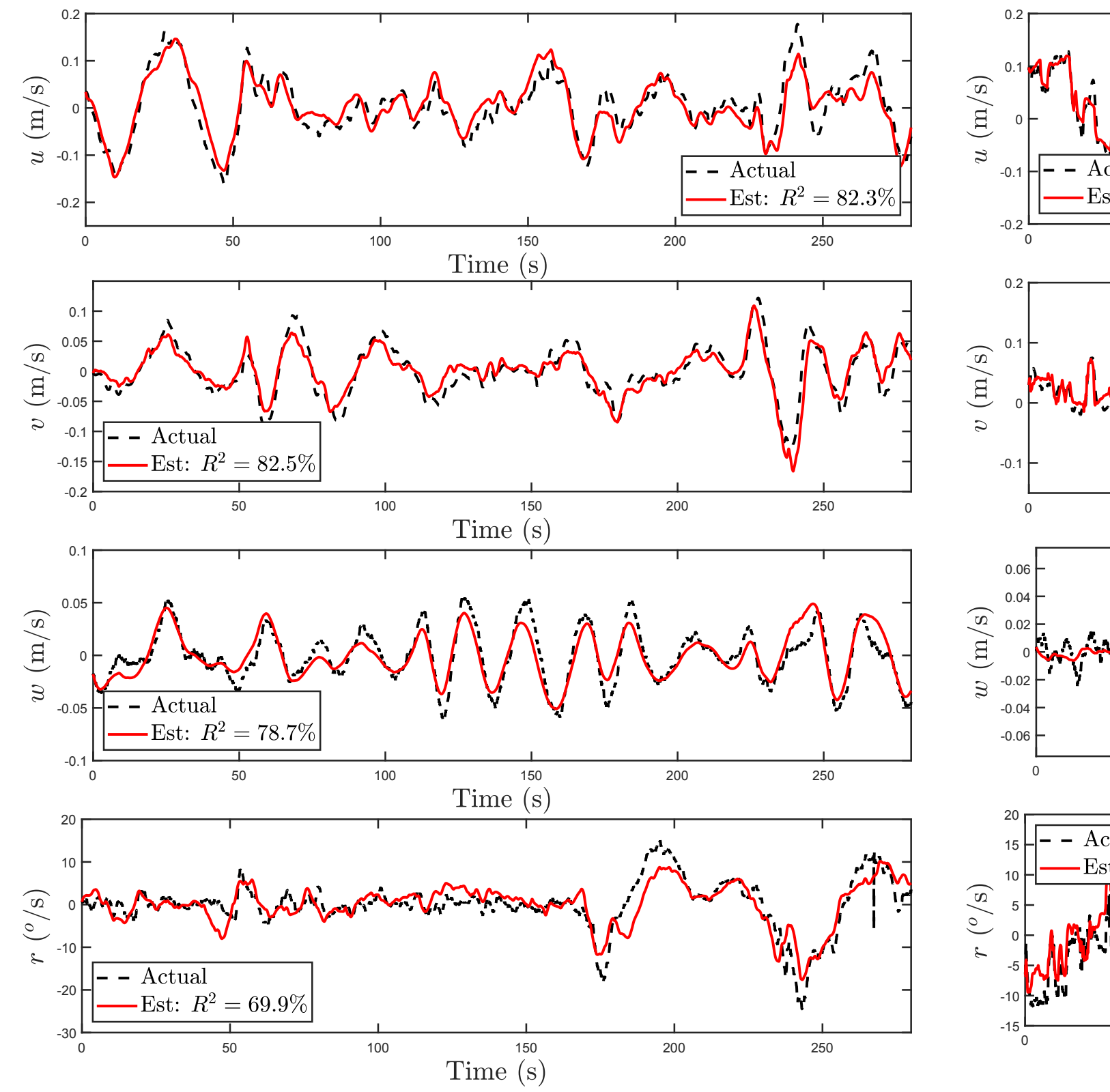

Fig. 7. Actual and simulated body velocities, with corresponding $R^2$ values, during a cross-validation experiment.

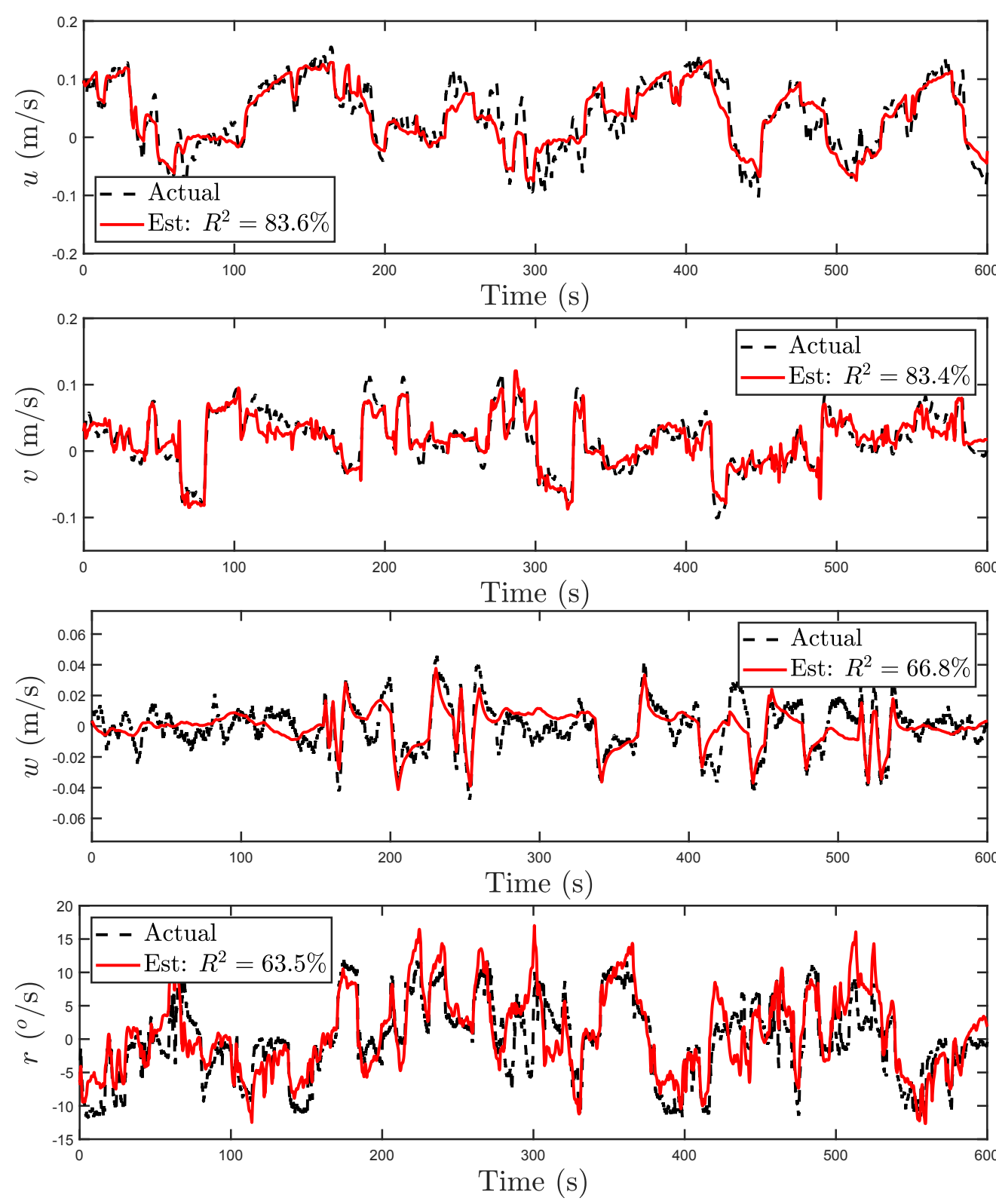

Fig. 8. Actual and simulated body velocities, with corresponding $R^2$ values for each DoF, during the teleoperation experiment.

### C. Cross-Validation

In this section, we evaluate the ability of the dynamic model, identified from the aforementioned experiment, to predict the vehicle's body velocities in experimental sessions with trajectories different from the one used for parameter identification. First, the generalizability of the identified dynamic parameters was assessed using 15 distinct reference trajectories of 280s, each generated by the proposed optimization algorithm and tracked by the vehicle with the PID controller. The simulated body velocities, obtained from the identified model, were compared against the measurements using the $R^2$ metric, as summarized in Fig. 6. Fig. 7 shows the simulated and actual velocities during an indicative experiment among the 15 trajectories. The results indicate that the identified model effectively generalizes to unseen trajectories and provides sufficiently accurate estimates of the body velocities, with the yaw rate consistently exhibiting the lowest $R^2$ values and the widest distribution, as expected.

To further assess the model, the identified parameters were tested in a more demanding and extended scenario involving abrupt velocity commands from joystick teleoperation, thereby imposing a significantly greater challenge on the dynamic model compared to the smooth trajectories generated using Bézier curves. Fig. 8 presents the simulated and actual velocities during a 10-minute teleoperation experiment. As shown, the velocities obtained from the forward propagation of the dynamic model closely match the measured data. This agreement confirms that the proposed methodology successfully generates excitation trajectories, enabling the identification of dynamic parameters that remain valid across a wider range of vehicle maneuvers, such as those encountered during teleoperation.

## V. CONCLUSIONS

In this paper, we presented a complete framework for the identification of the dynamic parameters governing the coupled 4-DoF motion of a UV. By exploiting Bézier curves, we generated a dynamically feasible and safe trajectory that effectively excites the system dynamics. The experimental results demonstrated that the resulting dynamic model, estimated through the optimized trajectory, is capable of reliably predicting the vehicle's velocity across a variety of maneuvers. The proposed methodology for generating excitation trajectories can be easily adapted to the operational requirements of other underwater platforms and workspace environments. Moreover, it solely entails robust state feedback and the presence of a conventional controller, e.g., PID, rendering it particularly suitable for modular UVs, whose payload configurations may frequently change.

Regarding our future work, we aim to extend the methodology to identify the full 6-DoF dynamic model of the UV, including roll and pitch dynamics, while incorporating visibility constraints to ensure the detection of an adequate number of ArUco markers. In addition, we plan to validate the estimated model in open-sea conditions. Furthermore, we intend to identify the dynamics of an Underwater Vehicle-Manipulator System (UVMS) with multiple manipulators and to investigate the effect of the cable on the vehicle's dynamics. Finally, we plan to integrate the identified model into a fault detection and diagnosis scheme to monitor the health of the thrusters and enable the detection and isolation of actuator failures.

## ACKNOWLEDGMENT

This work is supported in part by the NYUAD Center for Artificial Intelligence and Robotics, funded by Tamkeen under the Research Institute Award CG010.